\documentclass[journal,letter]{IEEEtran}

\usepackage{graphicx}

\usepackage[utf8]{inputenc}
\usepackage{amsmath}
\usepackage{hyperref}
\usepackage{bm}

\usepackage[percent]{overpic}

\usepackage{color}
\usepackage{xcolor,colortbl}

\usepackage{adjustbox}
\usepackage{array}

\usepackage{flushend} 

\newcolumntype{R}[2]{%
	>{\adjustbox{angle=#1,lap=\width-(#2)}\bgroup}%
	l%
	<{\egroup}%
}
\newcommand*\rot{\multicolumn{1}{R{60}{1em}}} 

\usepackage{cite}

\graphicspath{{./figs/}}

\definecolor{LightGreen}{rgb}{0.898039, 1, 0.8}

\newcommand{\Ex}{    \textcolor{blue}{({\bf Ex})} }
\newcommand{\G}{     \textcolor{teal}{({\bf G})} }
\newcommand{\R}{     \textcolor{cyan}{({\bf R})} }
\newcommand{\RG}{  \textcolor{orange}{({\bf RG})} }
\newcommand{\Sl}{     \textcolor{magenta}{({\bf S})} }
\newcommand{\inGM}{\textcolor{violet}{({\bf GM})} }

\newcommand{\PP}{PP}			
\newcommand{\LL}{LL}
\newcommand{\Lint}{L}
\newcommand{\Li}{L$_i$}

\newcommand{\PPi}{P$\Pi$}
\newcommand{\LPi}{L$\Pi$}
	
\newcommand{\PiPi}{$\Pi\Pi$}	
\newcommand{\PiPie}{$\Pi\Pi_e$}	
\newcommand{\LPie}{L$\Pi_e$}
\newcommand{\PPie}{P$\Pi_e$}
\newcommand{\PPPie}{sh 2P$\Pi_e$}		
\newcommand{\Pie}{$\Pi_e$}	
\newcommand{\Pe}{P$_e$}

\begin{document}

\title{\LARGE \bf A Grasping-centered Analysis for Cloth Manipulation}
\author{J\'ulia Borr\`as,~\IEEEmembership{Member,~IEEE,}  
Guillem Aleny\`a,~\IEEEmembership{Member,~IEEE,} 
Carme Torras,~\IEEEmembership{Fellow,~IEEE}      
	\thanks{The research leading to these results receives funding from the European Research Council (ERC) from the European Union Horizon 2020 Programme under grant agreement no. 741930 (CLOTHILDE: CLOTH manIpulation Learning from DEmonstrations) and  is also supported  by the Spanish State Research Agency through the María de Maeztu Seal of Excellence to IRI (MDM-2016-0656) and the “Ramon y Cajal” Fellowship RYC-2017-22703. }
	\thanks{The authors are  with Institut de Robòtica i Informàtica Industrial, CSIC-UPC,
		Llorens i Artigas 4-6, 08028 Barcelona, Spain. {\tt \{jborras, galenya, torras\}@iri.upc.edu}}%
}

\maketitle

\begin{abstract}
Compliant and soft hands have gained a lot of attention in the past decade because of their ability to adapt to the shape of the objects, increasing their effectiveness for grasping. However, when it comes to grasping highly flexible objects such as textiles, we face the dual problem: it is the object that will adapt to the shape of the hand or gripper. In this context,  the classic grasp analysis or grasping taxonomies are not suitable for describing textile objects grasps.
This work proposes a novel definition of textile object grasps that abstracts from
the robotic embodiment or hand shape and recovers concepts
from the early neuroscience literature on hand prehension skills. 
This framework enables us to identify what grasps have been used in  literature until now to perform robotic cloth manipulation, and allows for a precise definition of all the tasks that have been tackled in terms of manipulation primitives based on regrasps. In addition, we also review what grippers have been used.
Our analysis shows how the vast majority of cloth manipulations have relied only on one type of grasp, and at the same time we identify several tasks that need more variety of grasp types to be executed successfully. Our framework is generic, provides a classification of cloth manipulation primitives and can inspire gripper design and benchmark construction  for cloth manipulation.



\end{abstract}


\IEEEpeerreviewmaketitle

\section{Introduction}
\label{sec:introduction}


Robot manipulation in human environments has experienced great progress in recent years ~\cite{torras2016service}. However, efforts have been focused mostly on rigid objects, and core capabilities such as grasping, placing, or handing to a person still remain a hard and unsolved problem when dealing with challenging objects such as textiles. Indeed, textiles present many additional challenges with respect to rigid object manipulation, including difficult perception, complexity in modeling the object and predicting its behavior ~\cite{jimenez2017visual}, but it is of special importance in the case of domestic  and service robotics as textiles are present in many contexts of our daily lives. 

As a consequence, this topic is gaining a lot of attention from  EU Research committees. This work is framed in the ERC advanced grant CLOTHILDE, started in 2018, which aims to develop a theory of cloth manipulation based on  a topology-rich, compact cloth representation, advanced cloth perception, probabilistic motion planning and learning from demonstrations. In the project, we are studying how to define manipulation complexity and/or performance measures in terms of used grasps, computational time and quality of goal attainment. However, for that to be possible, we need a benchmark. 
Finding a proper benchmark and replicable experiments is challenging for robot manipulation in general ~\cite{Bonsignorio2015,quispe2018taxonomy} and, in particular for cloth-like objects, benchmarks are almost inexistent. 
We believe that a proper classification of the types of grasps and manipulation primitives used in handling textile objects is one of the steps necessary to develop a future such benchmark. 


In this work, we analyze and classify the types of grasps and the manipulation primitives that have appeared in literature, focusing on manipulations for handling clothes, including sorting, folding, unfolding, picking \& placing, hanging, etc. This is one of the three demanding applications addressed in the CLOTHILDE project, the other two being fitting elastic covers and helping handicapped people to dress, which we leave out of the scope of this paper.

 Despite being a relatively novel topic, there is already an extensive literature about robots perceiving and dealing with clothes in household tasks. A good recent review covering cloth-like objects, among others, can be found in ~\cite{sanchez2018robotic}. One of the main conclusions of the study is that manipulation skills still remain underdeveloped compared to vision or control for cloth manipulation and that improved end-effectors need to be developed to better handle this kind of objects. Indeed, there are only a few works focusing specifically on grasping of cloth-like objects, that is, on how and in how many different ways cloth can be grasped, and most of them have been developed for an industrial application and are not suitable for domestic environments ~\cite{Koustoumpardis2004,MOULIANITIS1999}. 

\begin{figure*}[t]
	\includegraphics[width=\textwidth]{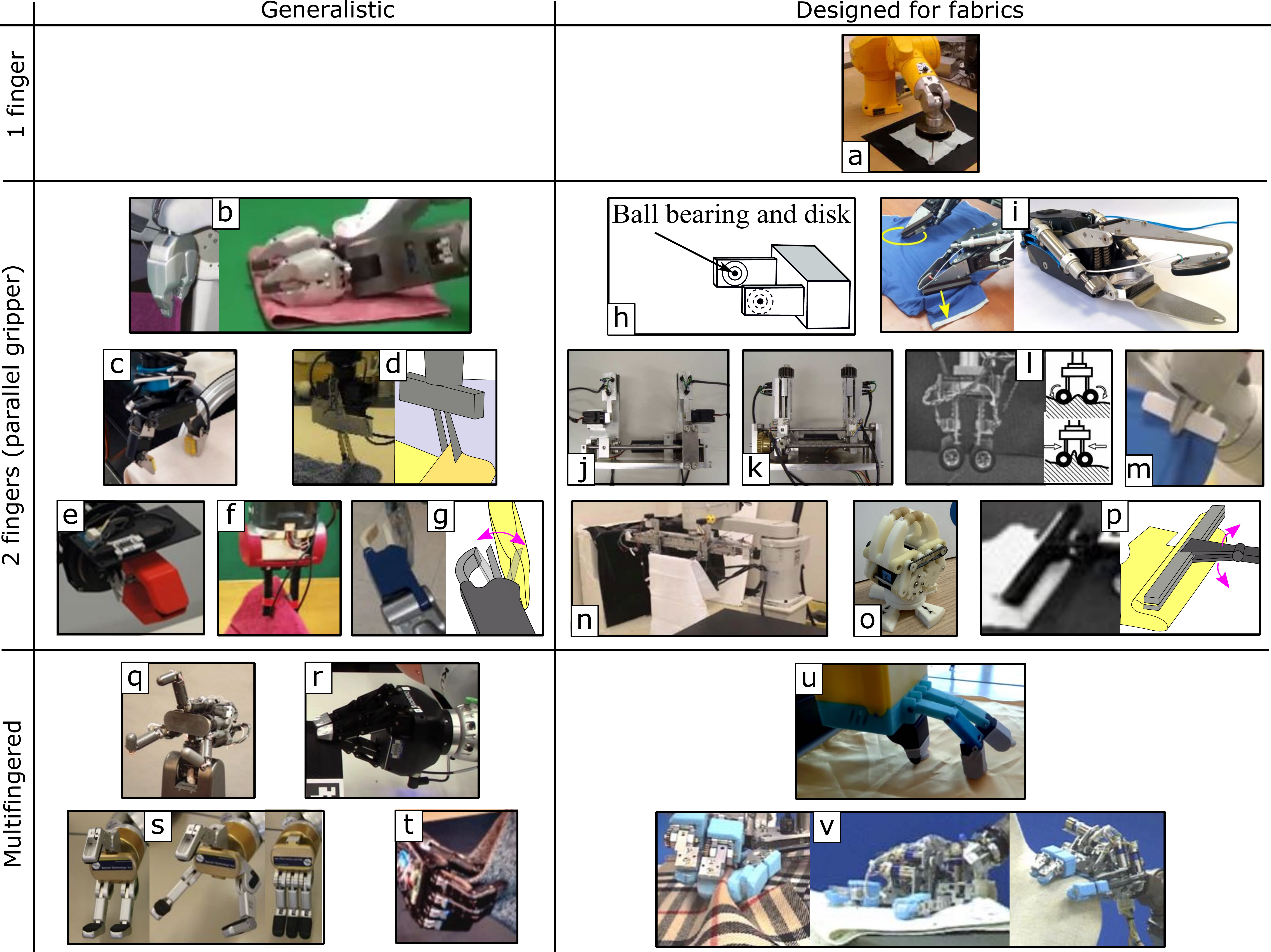}
	\centering
	\caption{
	 \textbf{a:} captured from ~\cite{cuen2008action}, 
	\textbf{b:}  PR2 gripper, image taken from ~\cite{maitin2010cloth}, 
	\textbf{c:} gripper used in ~\cite{yuba2017unfolding},
	\textbf{d:} Parallel gripper picture taken from ~\cite{willimon2011model}, 
	\textbf{e:} parallel gripper from the Nekonote robot arm from the RT Corporation used in ~\cite{moriya2018method},
	\textbf{f:} Baxter gripper, taken from ~\cite{li2015folding},	 
	\textbf{g:}  HRP2 gripper, from ~\cite{kita2011clothes}, 
	\textbf{h:}  Gripper from ~\cite{osawa2007unfolding},
	\textbf{i:}  Gripper used at the Clopema EU project ~\cite{Doumanoglou2016}, designed in ~\cite{Le2013development}, 
	\textbf{j:} and 
	\textbf{k:} taken from ~\cite{sahari2010edge}, 
	\textbf{l:} gripper with rolling fingertips introduced in ~\cite{kabaya1998service} and also used in ~\cite{hamajima2000planning}, 
	\textbf{m:} from ~\cite{jia2017manipulating}, 
	\textbf{n:}  taken from ~\cite{shibata2012fabric},
	\textbf{o:}  gripper used in ~\cite{colome2018dimensionality}, 
	\textbf{p:} image from ~\cite{Foldy2010}, 
	\textbf{q:} high speed 3 fingered hand from ~\cite{namiki2003development} used in ~\cite{yamakawa2011motion}, and
	\textbf{r:} RobotiQ 3-Finger Adaptive Robot Gripper, image from ~\cite{ruan2018accounting}, 
	\textbf{s:} Barret hand used in ~\cite{ramisa2012using,Monso2012,balaguer2011combining}, 
	 \textbf{t:} Shadow Dexterous Hand used in ~\cite{twardon2015interaction}, 
	\textbf{u:} underactuated gripper presented in ~\cite{koustoumpardis2014underactuated}, and
	\textbf{v:} 3 fingered hand and palm design for cloth manipulation in ~\cite{ono2001picking,ono2007better}.
	 }.
	\label{fig:grippers}
\end{figure*}

On the contrary, grasping has been studied extensively for rigid objects ~\cite{Feix2016Grasp,prattichizzo2008grasping}. 
However, most works assume not only that the object is rigid, but also that it is fixed inside the hand and can be held in any direction.  This is not true for clothes in almost any case. 
Furthermore, grasp classification has been highly dependent on object shape and size, while this distinction does not make much sense for clothes because cloth will conform to the shape of the gripper. 
Note also that rigid-object grasping research has started only recently to consider extrinsic contacts ~\cite{dafle2014extrinsic,Chavan-Dafle2015,Eppner2015}, that is, the use of contacts with the environment to improve grasp success. 
Instead, cloth manipulation has been utilizing extrinsic contacts 
since the very beginning, because they are badly needed to deal with this type of extremely deformable materials. It is worth highlighting that, in the context of rigid objects, environmental contacts are viewed as exploitable constraints to ease their grasping ~\cite{Eppner2015}, but not as grasps {\it per se} as they don't fix their pose. Instead, such contacts can be considered grasps for nonrigid objects, since they just partly confine their shape in the same way as grippers do. Finally, cloth manipulation is mostly bimanual and most of the grasps have to be used in couples.
In conclusion, we believe that further study of grasping centered on clothes, and considering all the aspects mentioned above, is necessary and could improve the robustness, autonomy and versatility of cloth manipulation by robots.

Another interesting insight of the review in ~\cite{sanchez2018robotic} is that the literature shows a clear tendency to develop techniques focused on a specific task and, thus, finding general solutions remains a major open issue. Identifying manipulation primitives that are common across different tasks can be of crucial relevance to move forward in this direction.

In this work we pave the way towards this end by analyzing previous attempts at robotizing specific cloth manipulation tasks under a common framework derived from three distinguishing traits of textiles: 1) their conforming to the gripper geometry (dual of soft manipulation), 2) the possibility to treat environmental contacts as grasps, thus allowing for a unified treatment, and 3) the need of two simultaneous grasps (bimanual handling) to accomplish most tasks. These traits permit abstracting from the particularities of existing grippers and defining some handy geometries and their combinations, inspired by the concepts of virtual fingers and opposition spaces from early neuroscience works on hand prehension ~\cite{iberall1986opposition}. The proposed framework delineates a well-defined landscape, which permits identifying grasps missing in previous research that may supply the versatility we are aiming at, leading to alternative ways of performing a given task or even the capacity to tackle new tasks. Other far-reaching possibilities are the design of grippers suitable to accomplish a given repertoire of tasks, and building benchmarks to assess the performance of perception, control and manipulation algorithms using some type of gripper to perform a precisely specified subset of tasks.


The paper is structured as follows.  \autoref{sec:grippers} analyzes all the grippers used for cloth manipulation. In \autoref{sec:framework} we propose a framework to characterize textile grasps and manipulations, which is then applied in \autoref{sec:analyzingManipulations} to systematize grasps and manipulations tackled in literature. In \autoref{sec:applications} we show an experiment to assess the influence of grasp type on task performance, and discuss the implications of our analysis and several possible applications. Finally, in \autoref{sec:conclusions} we provide conclusions and outline our future work.

\section{Gripper designs for manipulating textiles} \label{sec:grippers}

\autoref{fig:grippers} shows, as far as the authors know, all the grippers and robotic hands used in the literature focusing on cloth manipulation and presenting robotic experiments. Next we will describe  how these grippers are used and what kinds of grasps they can perform.

First, we observe that when cloth is manipulated by sliding a point contact on a table, no real gripper is needed (\autoref{fig:grippers}.\textbf{a}). Of course, then the functionality is limited to flattening clothes on a table. Most papers tackling general cloth manipulation use simple off-the-shelf parallel grippers like  \autoref{fig:grippers}. \textbf{b} to \textbf{g}, from different robot platforms like PR2 robot, HIRO,  Nekonote robot arm, Baxter and HRP2 robot, respectively. They all have relatively small planar fingertips and perform pinch grasps.  Others use generic robotic hands such as \autoref{fig:grippers}-\textbf{q} to \textbf{t}, corresponding to the RobotiQ, Barret and Shadow hands respectively. In these cases, the hands usually perform a pinch grasp between the thumb and 2 fingers, except in the case of ~\cite{ramisa2012using,Monso2012} that use different finger configurations from the Barrett hand to pick and place crumpled clothes, discussed in \autoref{tab:graspsInPapers}.

It is interesting to note how those tasks that require a bit more of support use grippers that perform a grasp between two lines. These are \autoref{fig:grippers}. \textbf{m}, \textbf{o} and \textbf{p}. The linear grip in \textbf{p} is used to grasp all the side of a T-shirt, therefore simplifying a bimanual folding operation to using a single hand. The gripper in  \autoref{fig:grippers}. \textbf{m} is used in ~\cite{jia2017manipulating} to maintain tension between grasps and therefore a larger part of the cloth can be controlled. Finally, the gripper in \autoref{fig:grippers}.\textbf{o} was used for the task in ~\cite{colome2018dimensionality}. Colomé and Torras apply learning techniques to fold a T-shirt  by rotating the shoulders of the T-shirt in the air. The gripper in the figure performs a grasp between two lines of very short length, and it could almost be considered as a pinch grasp. However, later they redesigned the gripper into the version shown at the top left of \autoref{fig:experimentSetup}. The grasp between longer lines  enables  to transmit more  torque effectively to the cloth, reducing the training time. A comparison of manipulation motions using different gripper geometries will be presented in \autoref{subsec:experiment}.

Finally, we will pay a bit more of attention to the grippers specially designed for manipulating clothes in domestic environments (\autoref{fig:grippers}.\textbf{h} to \textbf{p} and \textbf{u} and \textbf{v}, that is, the right column of the table).
Some of these are parallel grippers with some modification at the fingertips, like \autoref{fig:grippers}.\textbf{h} that has a ball bearing to ease sliding, or \autoref{fig:grippers}.\textbf{k} and \textbf{l} that use rolling fingertips to allow sliding along or pulling upwards a pinched cloth, respectively. \autoref{fig:grippers}.\textbf{i} is effectively a parallel gripper with a very thin finger to slide under flat clothes, and a touch sensor at the other fingertip. A linear brush was added at the side to be able to perform a flattening operation by sliding it on top of the cloth, as shown in the left image in \autoref{fig:grippers}.\textbf{i}. Designs \textbf{j} and \textbf{n} show a similar idea: to have small grippers at the fingertips of a parallel gripper in \textbf{j} or joined by a prismatic actuator in \textbf{n}. These designs allow in-hand manipulation such as grasp gaiting, or edge tracing with a single gripper in the case of \textbf{n}, although such a big long gripper would not be practical in domestic environments. \autoref{fig:grippers}.\textbf{u} shows an under-actuated gripper specially designed for pinching and clamping edges and corners of a flat cloth laying on a table. In addition, it could perform the same pinch grasp as the multifingered robot hands that we mentioned before. Finally, the hand in \autoref{fig:grippers}.\textbf{v} is a humanoid-like 3-fingered hand with a palm, and it is specially designed to be able to slide on a cloth with a flat hand, perform a pinch between the palm and one finger or a pinch between two fingers. These operations are identified as requirements for cloth manipulation in ~\cite{ono2001picking}.

In conclusion, off-the-shelf grippers can only perform pinch grasps, which suffice to execute most of the tasks. Other specially-designed grippers  exhibit a higher versatility, but more testing is needed to validate their usefulness for a wide range of cloth manipulation tasks.

\section{A framework proposal for robotized clothes handling} \label{sec:framework}
\subsection{Characterizing grasps in terms of opposing geometric entities} \label{subsec:grasping_framework}
 
Grasping analysis and taxonomies have been highly influenced by neuroscience works in the 80's that defined parameters to characterize hand prehensile postures in terms of virtual fingers and opposition spaces. Iberall's works ~\cite{iberall1986opposition,arbib1985coordinated} established that, to grasp an object, at least two opposing forces against the object surface are needed. 
The opposition space is defined by the possible directions of the two forces. 
The concept of virtual finger (VF) groups fingers or hand surfaces that apply forces in the same direction into functional units. In other words, each finger and hand surfaces working in a grasp are assigned to one of the VF, and most grasps can be defined by just two VFs exerting opposing forces. This permits the abstraction from the mechanics of the hand and its fingers. 

\begin{figure*}[t!]
	\centering
	\includegraphics[width=0.75\textwidth]{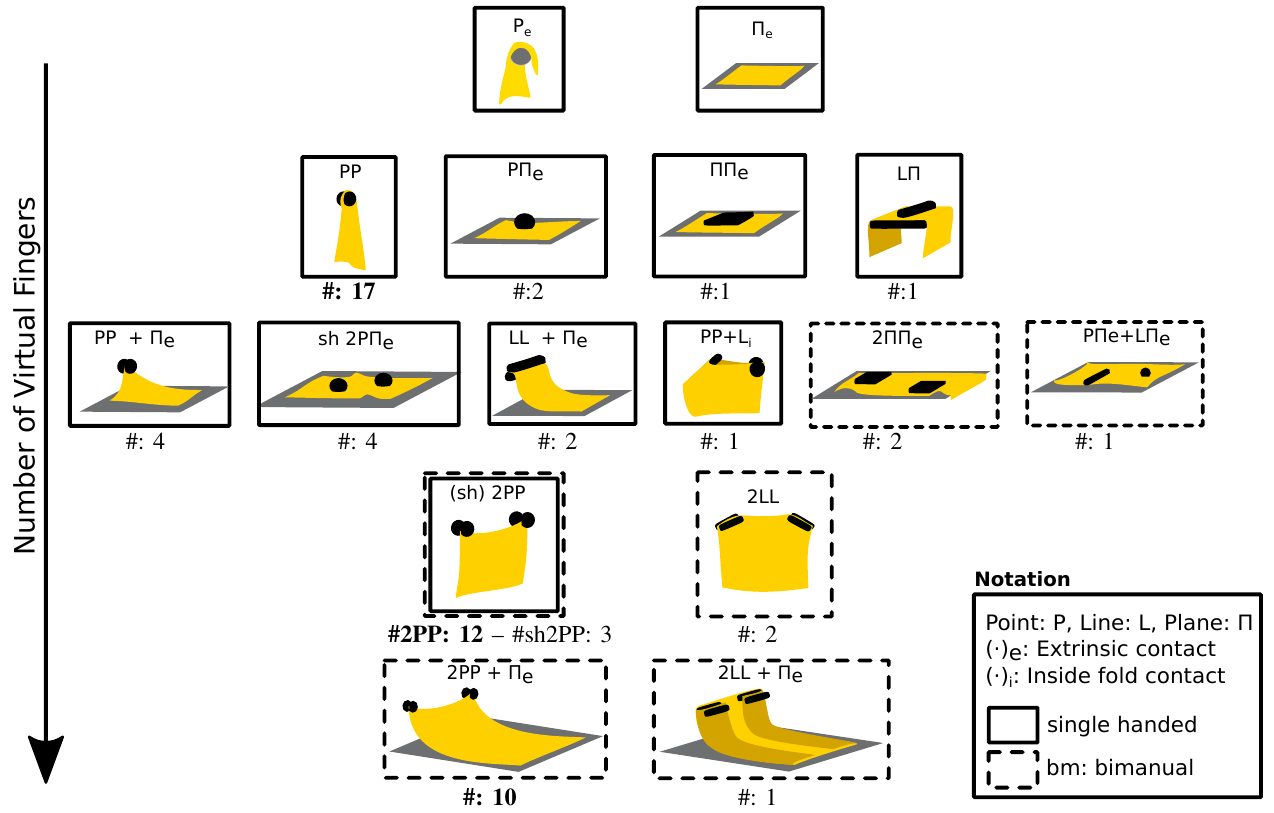}
	\caption{Graphical representation of the grasps appearing in literature, detailed by citation in \autoref{tab:graspsInPapers}. The numbers correspond to the times the corresponding grasp has been used in literature.}
	\label{fig:taxonomy}
\end{figure*}

Iberall's works were centered on the human hand, and they described hand postures as collections of VFs able to apply effective forces against the object. She showed that standard prehensile classifications like Cutkosky's ~\cite{cutkosky1989grasp} could be captured with this framework and classified in terms of VFs and opposition directions, as done later by Feix et al. ~\cite{Feix2016Grasp}.

Although traditionally the concept of VF has been applied to the human hand, the framework could be much widely used to characterize grasping by artificial means as well ~\cite{iberall1997human}. Indeed, it is shown in ~\cite{iberall1991parameterizing} that a grasp could be fully described and analyzed in terms of the parameters that describe VFs: parameters of the grasping contact surface patch (length, width and orientation w.r.t. the palm), the orientation of the force the VF could exert and the sensor information. 

\begin{table}[tb]
	\caption{Grasps used in literature}
	\label{tab:graspsInPapers}
	\centering
	\setlength\tabcolsep{1pt} 
	\renewcommand{\arraystretch}{1.15} 
	\begin{tabular}{|l|c|c|}
	\hline
	Reference & \begin{tabular}{c}Gripper\\ in \autoref{fig:grippers} \end{tabular}& Grasps used \\
\hline
	{\scriptsize Cu{\'e}n Roch{\'\i}n et al. 2008	} \cite{cuen2008action}					  	&	a				& \PPie 									\\

	{\scriptsize Maitin-Shepard et al. 2010	} \cite{maitin2010cloth}						    &	b				&  \PP, 2\PP, 2\PiPie, 2\PP+\Pie 			\\
	{\scriptsize Bersch et al. 2011			} \cite{bersch2011bimanual}						    &	b				&  \PP, 2\PP, 2\PiPie						\\
	{\scriptsize Miller et al. 2012           }\cite{miller2012geometric, 
									cusumano2011bringing,miller2011parametrized,
										van2010gravity, lakshmanan2013constraint}		        &	b				& \PP+\Pie, 2\PP+\Pie 					\\
	{\scriptsize Lee at al. 2015				}\cite{lee2015learning}							&	b 				& 2\PP, 2\PP+\Pie 							\\

	{\scriptsize Yuba et al. 2017				} \cite{yuba2017unfolding}						&	c				& \PP, 2\PP, 2\PP+\Pie 						\\
	{\scriptsize Koishihara et al. 2017			} \cite{koishihara2017hanging}		  	  	    &	c				& \PP, \PP+\Li$^1$							\\
	{\scriptsize Willimon et al. 2011				}  \cite{willimon2011model}				  	&	d				&  \PP+\Pie									\\  
	{\scriptsize Moriya et al. 2018				}  \cite{moriya2018method}				  	    &	e				&  \PP           							\\  
	{\scriptsize Li et al. 2015				} \cite{li2015regrasping,li2015folding}			    &	f				&  \PP, 2\PP, 2\PP+\Pie, \PP+\Pie 			\\
	{\scriptsize Kita et al. 2011				} \cite{kita2011clothes}						&	g	 			&  \PP, 2\PP, 								\\
	{\scriptsize Osawa et al. 2007			} \cite{osawa2007unfolding}						    &	h	 			&  \PP, 2\PP, 								\\ 
	{\scriptsize Doumanoglou et al. 2016		} \cite{Doumanoglou2016}					    &	i				&  \PP, 2\PP, 2\PP+\Pie, \PPie+\LPie 		\\
	{\scriptsize Sun et al. 2015					} \cite{sun2015accurate}				  	&	i				&  \PP+\Pie, 2\PP+\Pie 						\\
	{\scriptsize Sahari et al. 2010			} \cite{sahari2010Clothes,  
											sahari2010edge,salleh2008inchworm}			        &	j,k				&  \PP, 2\PP, sh 2\PP 						\\
	{\scriptsize Hamajima et al. 2000				} \cite{hamajima2000planning}			  	&	l				& \PP, sh 2\PPie							\\
    {\scriptsize Kabaya et al. 1998				} \cite{kabaya1998service}			  	  	    &	l				& \PP, sh 2\PPie 							\\
	{\scriptsize Jia et al. 2017				} \cite{jia2017manipulating}					&	m				& 2\LL, \LL+\Pie							\\
	{\scriptsize Shibata et al. 2012			} \cite{shibata2009wiping,shibata2012fabric}	&	n				&  sh 2\PP, sh 2\PP+\Pie 					\\
	{\scriptsize Colome and Torras 2018		}  \cite{colome2018dimensionality} 				    &	o				& 2\LL, 2\LL+\Pie							\\

	{\scriptsize Sugiura et al. 2010			} \cite{Foldy2010}								&	p				& \LL+\Pie 									\\
	{\scriptsize Yamakawa et al. 2011			} \cite{yamakawa2011motion}					  	&	q				& 2\PP										\\
	{\scriptsize Ruan et al. 2018			} \cite{ruan2018accounting}					  	    &   r				& 2\PP+\Pie									\\
	{\scriptsize Balaguer and Carpin 2011		} \cite{balaguer2011combining}				  	&	s				& 2\PP, 2\PP+\Pie 							\\
	{\scriptsize Monso et al. 2012				} \cite{Monso2012}						  	    &	s				& \PP, sh2\PP, \LPi $^2$ 					\\ 
	{\scriptsize Ramisa et al. 2012				}  \cite{ramisa2012using}			  	  	    &	s				& \PP									    \\
	{\scriptsize Twardon et al. 2015				} \cite{twardon2015interaction} 		  	&	t				& \PP, 2\PP, 								\\
	{\scriptsize Koustoumpardis et al. 2014		} \cite{koustoumpardis2014underactuated,  	
																Koustoumpardis2017}	  	        &	u				& \PP, sh 2\PPie 							\\	
	{\scriptsize Ono and Takase 2007				} \cite{ono2007better,ono2001picking}	  	&	v				& \PP, sh 2\PPie, \PPie, \PiPie 			\\
\hline
Total grasp instances 																	& 											& 63 										\\ 	
\hline
	\end{tabular}
\flushleft
	Notes\\
	$1.$ Line contact is achieved by grasping a hanger with the second gripper.\\
	$2.$  From the finger configurations in \autoref{fig:grippers}-{\bf s}, the far left configuration leads to a sh2\PP{}, where the first \PP{} is between one finger and the right side of thumb and the second \PP{} is achieved with the other finger and the left side of the thumb. For the next to the right image, when fingers close in symmetric configuration, the achieved grasp is a pinch that is either a \PP{} grasp or a sh2\PP{} grasp as before. With the configuration on the right, the hand achieves a \LPi{} grasp where the line is formed with the 3 aligned fingertips that will exert force against the plane of the palm. All grasps are used to pick crumpled clothes.

\end{table}

We propose to define textile grasps in terms of the number and geometry of VFs. In other words,  {\em geometric virtual fingers} define the shape of the contact patch and the orientation with respect to each VF. Implicitly, this also determines the possible directions of the opposition forces. For textiles, we can characterize most of the grasps using only three geometric entities: points (P), lines (L) and planes ($\Pi$); and \textit{we define a grasp as a set of opposing geometric entities}. We believe this definition makes sense in the context of pure shapeless objects that will conform to the geometry of the grasping device, also implying the shape of the object will not influence the definition of the grasp.

\begin{table*}[t!]
	\caption{Description of tasks}
	\label{tab:tasks}
\begin{center}
\setlength\tabcolsep{3pt} 
\renewcommand{\arraystretch}{1.15} 
	\begin{tabular}{|l|lll|}
\hline
\textbf{Task Name} &  \textbf{Start} 		&{\bf Steps} 																																																							&  \textbf{End}  \\ \hline \hline

1. (a) Unfold in the air 			&	(H1) \PP{} at p			&	\begin{tabular} {l l}
														${\bf 1)}$ (H2) \RG Gr v: \PP{} 	&										
														${\bf 2)}$ (H1) \RG Rlse \PP{} 	\\			
														${\bf 3)}$ (H1) \RG Gr v: \PP{} 		&									
													\end{tabular}														&		2\PP{} unfd	\\ \hline
1. (b) Unfold in the air 			&	(H1) \PP{} at p			&	\begin{tabular} {l l}
														${\bf 1)}$ (H1) \Ex Add \Pie: \PP{}+\Pie 	&										
														${\bf 2)}$ (H1) \Sl Slide to reveal corner  	\\			
														${\bf 3)}$ (H2) \RG Gr v: 2\PP{}+\Pie 		& ${\bf 4}$	\Ex Rm \Pie: 2\PP{} \\
														${\bf 5)}$ Repeat from ${\bf 1)}$ to ${\bf 4)}$ changing hands 
													\end{tabular}														&		2\PP{} unfd	\\ \hline

1. (c) Unfold in the air tracing		&	 \Pie{} crumpled &	\begin{tabular} {l l}
														${\bf 1)}$ (H1) \G Gr v: \PP{} +\Pie &
														${\bf 2}$ (H1) \Ex Rm \Pie: \PP{} \\
														${\bf 3)}$ (H2)\RG Gr e near H1: 2\PP{} &			  ${\bf 4)}$ \Sl Trace edge until H2 finds v 											
													\end{tabular}														&		2\PP{} unfd	\\ \hline
2. Fold on table			&	\Pie{} flat or fd	    &	\begin{tabular} {l l}
														${\bf 1)}$ \G Gr 2 v: 2\PP{} + \Pie	& ${\bf 2)}$ \inGM  Fold						
														${\bf 3)}$ \R Rlse 2\PP{} on tbl										
													\end{tabular}														&		\Pie{}  fd	\\ \hline
3. (a) Fold in the air with table  	&	2\LL{} unfd	&	 \begin{tabular} {l l }
														${\bf 1)}$ \inGM Pronation rot. each H	&	${\bf 2)}$ \Ex Add \Pie: 2\LL+\Pie{} folding	\\	
														${\bf 3)}$ \R Rlse 2\LL{} from inside fold  &							
														\end{tabular}													&		\Pie{}  fd	\\ \hline
3. (b) Fold in the air dynamic  	&	2\PP{} unfd	&	 \begin{tabular} {l l }
														${\bf 1)}$ \inGM Fast move cloth until bottom vertices move up	&	\\	
														${\bf 2)}$ \RG Regrasp  2\PP{} to add the bottom vertexs  &							
														\end{tabular}													&		2\PP{} fd \\ \hline														
4. Place on table flat or 
folding							&	2\PP{} unfd	&  \begin{tabular} {l l}
														${\bf 1)}$ \Ex Add \Pie: 2\PP{} + \Pie	 &					
														${\bf 2)}$ \R Rlse 2\PP{} on tbl										
													\end{tabular}														&		\Pie{} flat or fd	\\ \hline
5. (a) Flatten cloth			&	\Pie{}	p.flat			&	\begin{tabular} {l }
														${\bf 1)}$ \G Gr 2\{\LPie , \PPie\} \\				
														${\bf 2)}$ \Sl Slide H1, hold H2												
													\end{tabular}														&		\Pie	(p)flat \\ \hline
5. (b) Flatten cloth			&	\Pie{}	p.flat			&	\begin{tabular} {l }
														${\bf 1)}$ \G Gr 2\PiPie{} \\			
														${\bf 2)}$ \Sl Slide H1 \& H2 in opp. dir.										
													\end{tabular}														&		\Pie{} (p)flat	\\ \hline
5. (c) Flatten cloth			&	\Pie{}	p.flat			&	\begin{tabular} {l l}
														${\bf 1)}$ \G Gr \{v, e\}: $\{1, 2\}$\PP+\Pie{} \\	   		
														${\bf 2)}$ \inGM Pull H(s) to flatten &${\bf 3)}$ \R Rlse	 $\{1, 2\}$\PP on table										
													\end{tabular}														&		\Pie	(p)flat\\ \hline
5. (d) Flatten cloth			&	\begin{tabular}{l}
									\Pie	p.flat \\
									(1 v folded)
									\end{tabular}	&	\begin{tabular} {l l}
														${\bf 1)}$ (H1) \G Gr p \PPPie{} to hold cloth & ${\bf 2)}$ (H2) \RG Gr v: \PP{} while H1 holds	\\				
														${\bf 3)}$ (H1) \R Rlse \PPPie{} 	& ${\bf 4)}$ (H2) \inGM Unfold v and rls \PP{} 								
													\end{tabular}														&		\Pie{} (p)flat	\\ \hline
6. Separate from pile (pick \& place)	&	\Pie{} crumpled	&	\begin{tabular} {l l}
														${\bf 1)}$ \G Gr p w. \PP{} or \PiPi{}	&	${\bf 2)}$ \R Rlse 					
													\end{tabular}														&		 \Pie{} crumpled	\\ \hline
7. Pick \& place folded cloth	&	\Pie{} fd				&	\begin{tabular} {l l}
														${\bf 1)}$ \G Gr  2\PP{} from 2 sides and lift	&							
														${\bf 2)}$ \R Rlse 2\PP{} on table											
													\end{tabular}														&		\Pie{} fd	\\ \hline
8. Clamp (p)flat cloth (pick up) 	&	\Pie{} (p)flat		&	\begin{tabular} {l l}
														${\bf 1)}$ \inGM Move gripper finger under cloth \\						
														${\bf 2)}$ \G Grasp cloth at e/v: \PP{} + \Pie{} & ${\bf 3)}$ \Ex Rm \Pie{}  												
													\end{tabular}														&		\PP{} at e/v	\\ \hline

9. Pinch (p)flat cloth (pick up)	&	\Pie{} (p)flat	&	\begin{tabular} {l l}
														${\bf 1)}$ \G Gr \PPPie{} & ${\bf 2)}$ \inGM Bring P contacts together \\
														${\bf 3)}$ \RG Rgr to \PP{} + \Pie		&  	${\bf 4)}$ \Ex Rm \Pie										
													\end{tabular}														&		\PP{} at p	\\ \hline
10. Pick single layer from stack	&	\Pie{} flat	&	\begin{tabular} {l }
														${\bf 1)}$ (H1) \G Gr p \PPie{} near edge 		\\
														${\bf 2)}$ (H1) \inGM Push and slide inwards cloth to create crease and hold		\\		
														${\bf 3)}$ (H2) \RG Grasp edge \PP{} on crease and \R Rlse H1	
													\end{tabular}														&	  \PP{} at p	\\ \hline

11. Grasp gaiting				&	(H1) \PP{} at p			&	\begin{tabular} {l l}
														${\bf 1)}$ (H2) \RG Gr p \PP{} near H1 	&									
														${\bf 2)}$ (H1) \RG Rlse \PP{}  		\\		
														${\bf 3)}$ Iterate					&							
													\end{tabular}														&	  \PP{} at p	\\ \hline
12. Hang 						& 2\PP{} unfd		&	\begin{tabular} {l l}
														${\bf 1)}$ \Ex Add \Pe: 2\PP{} + \Pe &								
														${\bf 2)}$ \R Rlse 2\PP{}
													\end{tabular}														&		\Pe	\\ \hline
13. Put hanger in T-shirt       & (H1) \PP{} at p      &	\begin{tabular} {l }
														${\bf 1)}$ (H2) \RG Insert side of hanger in neck hole, adding a \Lint{}  	\\						
														${\bf 2)}$ (H2) \inGM Pull between \Lint{} and \PP{} to insert other side of hanger \\ 
													\end{tabular}														&		\Lint{}\\ \hline
14.Japanese style T-shirt fold  & \Pie{} flat				&	\begin{tabular}{l}
														${\bf 1)}$ \G Gr$^2$: 2\PP{} + \Pie{}	         \\
														${\bf 2)}$ \inGM H1 moves to opposite edg, H2 holds	\\
														${\bf 3)}$ (H1) \RG Add edge p to \PP{} \hspace{2mm}
														${\bf 4)}$ \Ex Rm \Pie{} \\ 
														${\bf 5)}$ \inGM shake$^3$	 \hspace{2mm}
														${\bf 6)}$ \Ex Add \Pie{}  \hspace{2mm}
														${\bf 7)}$ \R Rlse 2\PP{} on tbl 								
													\end{tabular}														&	\Pie{} fd	\\ \hline		

\end{tabular}
\end{center}

\vspace{0.1cm}
Notation:   v: vertex/corner of cloth, e: point on the edge of cloth, p: any point interior of cloth\\
Abbreviations: H: hand, F: fold, Fd: Folded, Unf: unfold, Unfd: Unfolded, Gr: Grasp, Rm: Remove,
edg: edge, Rlse: Release, Tr: Trace, sh: single hand, bm: bimanual, p.flat: partially flat.
\end{table*}


We can easily see at this point that the environment geometries play a very relevant role as grasping agents to increase functionalities of the grippers. Therefore, we will consider environmental contacts as extra fingers, so that a geometric VF can be either intrinsic, that is, defined by the contact patch of the gripper/hand, or extrinsic, that is, defined by contact surfaces from the environment. Any extrinsic VF will be denoted with the subindex "$_e$". In this context, and as done also in ~\cite{dafle2014extrinsic}, we consider as a grasp an object lying on a table. That is, the table is an extrinsic planar VF, denoted \Pie{}.

Because cloth manipulation is intrinsically bimanual, we will denote bimanual grasps with the addition symbol, in a way that the flattening operation in \autoref{fig:grippers}.\textbf{i} is represented by \PPie{}+\LPie{}, and if the grasps are symmetric in both hands, like for instance two hands holding a cloth in the air with two pinch grasps, we simply write 2\PP{}.  We assume that any double grasp is bimanual, and we mark it as a single handed one otherwise, using "sh". For instance, the grasp in \autoref{fig:grippers}.\textbf{n} is defined as sh~2\PP{}. 
Similarly, a single grasp is assumed to be single handed, but can be marked as bimanual using "bm" if it is performed by two hands.

In the context of our framework, we define both prehensile and non-prehensile grasps. Any combination of geometries of different dimensions will lead to a non-prehensile grasp, while the same dimensions lead to prehensile ones. In general, according to definitions in literature ~\cite{bullock2012hand}, even when the two geometries are the same, like a \PP{} grasp or a \PiPi{} grasp, they perform a non-prehensile grasp because generally they will not fix the position of the entire object with respect to the hand. However, in the context of our framework, we will consider these as prehensile grasps (prehensile inside the limits of the hand), to distinguish them from the other "less" prehensile grasps. Finally, any grasp involving only extrinsic geometries, like \Pie{}, is also non-prehensile because it assumes the second opposition force is gravity.

In \autoref{subsec:applicationOnGrasps} we will apply this framework to describe all the grasps used by the grippers in \autoref{fig:grippers}, summarized and classified in \autoref{fig:taxonomy}.

\subsection{Manipulation primitives as grasp transitions}\label{subsec:grasping_manipulation}

Similarly as  in ~\cite{dafle2014extrinsic}, after having established the repertoire of grasps we will deal with, we can consider as manipulation primitives either regrasping actions or motions with a fixed grasp. Within this framework, we define six types of manipulation primitives:

\begin{itemize}
	\item \Ex Adding/removing extrinsic contacts. These are regrasp actions where the origin/destination has an extrinsic VF while the destination/origin doesn't, respectively.
	\item \G Grasping. A regrasp from a single extrinsic VF grasp (first row in \autoref{fig:taxonomy}) to any other grasp in the figure.
	\item  \R Releasing. The opposite of grasping, that is, a regrasp  from any grasp to a single extrinsic VF grasp.
	\item \RG Regrasping. Pure regrasp action, that is, any transition between grasps that does not belong to the two previous types.
	\item \Sl Sliding. Manipulation where the initial and end grasps are the same, but the cloth has moved with respect to the grasp without losing contact completely. Examples include edge tracing or flattening on a table.
	\item \inGM In-grasping motions. Single hand or bimanual grasps that are fixed and moved by the arms without performing any regrasp but changing the state of the cloth. Examples include applying tension between two \PP{} grasps or the motion to fold a cloth after is grasped.
\end{itemize}

Then, we can define a task as a sequence of these types of manipulations. In \autoref{subsec:applicationOnTasks}  we will describe all the tasks addressed in literature as sequences of manipulation primitives of the above six types.

\section{Analyzing cloth manipulation tasks using the proposed framework} \label{sec:analyzingManipulations}

\subsection{Gripper capabilities in terms of the proposed framework} \label{subsec:applicationOnGrasps}
The use of geometric virtual fingers is general in the sense that it can be applied to the human hand by mapping its five fingers and palm to a geometric arrangement of VFs, but it can also be applied to other generic grippers like the ones listed in the previous section. Revisiting the grippers in \autoref{fig:grippers} we realize that if we define a point as a relatively small flat surface, most of the grippers above  perform a \PP{} grasp (a pinch) (\autoref{fig:grippers}.\textbf{b}-\textbf{g}, \textbf{q}-\textbf{t}, \textbf{i} and \textbf{u}) or a sh 2\PP{} grasp (\autoref{fig:grippers}.\textbf{j} and \textbf{n}). Some of them perform a \PP{} grasp with additional abilities to roll over clothes (\autoref{fig:grippers}.\textbf{h, k} and \textbf{l}). In addition, the grippers in \autoref{fig:grippers}.\textbf{m} and \textbf{p} perform an \LL{} grasp.

\begin{table*}
	\caption{Distribution of tasks in literature}
	\label{tab:tasksInPapers}
	\centering 
\setlength\tabcolsep{2pt} 
	\begin{tabular}{| l |r|c|c|c|c|c|c|c|c|c|c|c|c|c|c|c|c|c|c|c|c|c|c|c|c|c|c|c|c|c|c|c|}
	\multicolumn{1}{l} {\textbf{Tasks}} & \multicolumn{1}{r}{}
	&\rot{{\scriptsize Osawa et al. 2007			}  \cite{osawa2007unfolding}} 
	&\rot{{\scriptsize Maitin-Shepard et al. 2010	} \cite{maitin2010cloth}} 
	&\rot{{\scriptsize Kita et al. 2011				}   \cite{kita2011clothes}} 
	&\rot{{\scriptsize Bersch et al. 2011			} \cite{bersch2011bimanual}} 
	&\rot{{\scriptsize Li et al. 2015				} \cite{li2015regrasping,li2015folding}}
	&\rot{{\scriptsize Doumanoglou et al. 2016		} \cite{Doumanoglou2016}} 
	&\rot{{\scriptsize Customano-Towner et al. 2011	} \cite{cusumano2011bringing}} 
	&\rot{{\scriptsize Sahari et al. 2010			} \cite{sahari2010Clothes,sahari2010edge,salleh2008inchworm}} 
	&\rot{{\scriptsize Shibata et al. 2012			} \cite{shibata2009wiping,shibata2012fabric}} 
	&\rot{{\scriptsize Yuba et al. 2017				} \cite{yuba2017unfolding}}
	&\rot{{\scriptsize Sugiura et al. 2010			} \cite{Foldy2010}}
	&\rot{{\scriptsize Miller et al. 2012			}\cite{miller2012geometric,miller2011parametrized,van2010gravity, lakshmanan2013constraint}} 	
	&\rot{{\scriptsize Lee at al. 2015				}\cite{lee2015learning}}
	&\rot{{\scriptsize Jia et al. 2017				} \cite{jia2017manipulating}} 
	&\rot{{\scriptsize Colome and Torras 2018		}  \cite{colome2018dimensionality} } 
	&\rot{{\scriptsize Yamakawa et al. 2011			} \cite{yamakawa2011motion}}
	&\rot{{\scriptsize Balaguer and Carpin 2011		} \cite{balaguer2011combining}} 
	&\rot{{\scriptsize Ruan et al. 2018				} \cite{ruan2018accounting}} 
	&\rot{{\scriptsize Cu{\'e}n Roch{\'\i}n et al. 2008	} \cite{cuen2008action}}
	&\rot{{\scriptsize Willimon et al. 2011				} \cite{willimon2011model}}  
	&\rot{{\scriptsize Sun et al. 2015					} \cite{sun2015accurate}} 
	&\rot{{\scriptsize Hamajima et al. 2000				} \cite{hamajima2000planning}} 
	&\rot{{\scriptsize Monso et al. 2012				} \cite{Monso2012}}
	&\rot{{\scriptsize Ramisa et al. 2012				} \cite{ramisa2012using}} 
	&\rot{{\scriptsize Moriya et al. 2018				} \cite{moriya2018method}} 
	&\rot{{\scriptsize Kabaya et al. 1998				} \cite{kabaya1998service}} 
	&\rot{{\scriptsize Koustoumpardis et al. 2014		} \cite{koustoumpardis2014underactuated, Koustoumpardis2017}} 	
	&\rot{{\scriptsize Ono and Takase 2007				} \cite{ono2007better,ono2001picking}}
	&\rot{{\scriptsize Twardon et al. 2015				} \cite{twardon2015interaction} } 
	&\rot{{\scriptsize Koishihara et al. 2017			} \cite{koishihara2017hanging}}
	&\rot{{\scriptsize Bell et al. 2010					} \cite{bell2010flexible}} 	    
 \\ \hline						
\rowcolor{LightGreen}																																													
	1. (a) Unfold in the air	&  6&x			 &x						&x			&x				&x 			&	x				&						&				& 				&   		&  			  &  			&  					&				& 				&				& 				&			& 					&					&				&                   &				&				&						 &				    & 						&				&				&					&			\\ \hline																																																																																																																																										
	1. (b) Unfold in the air	&  1&			 &						&			&				& 			&					&			x			&				& 				&   		&  			  &  			&  					&				& 				&				& 				&			& 					&					&				&                   &				&				&						 &				    & 						&				&				&					&			\\ \hline	\rowcolor{LightGreen}																																																																																																																																									
	1. (c) Unfold sliding	 	&  3&			 &						&			&				&			&					&						&	x			& 	x			&   x		&  			  & 			& 					&				& 				&				& 				&			& 					&					&				&                   &				&				&						 &				    & 						&				&				&					&			\\ \hline
	2. Fold on table			&  8&			 &x$^1$					&			&				&x			&	x				&						&				& 				& 	 		&  	x		  &  x			&  	x$^3$			&	x			& 				&				& 				&			& 					&					&				&                   &				&				&						 &				    & 						&				&				&					&			\\ \hline  \rowcolor{LightGreen}																																																																																																																																								
	3. Fold in the air 			&  1&			 &						&			&x$^1$			&			&					&						&				& 				&  			&			  & 			& 					&				& 	x			&				& 				&			& 					&					&				&                   &				&				&						 &				    & 						&				&				&					&			\\ \hline
	3. Dynamic fold 			&  1&			 &						&			&				& 			&					&						&				& 				& 	 		&  			  & 			&  					&				& 				&		x		& 				& 			& 					&					&				&                  	&				&				&						 &				    & 						&				&	 			&					&			\\ \hline	\rowcolor{LightGreen}																																																																																																																																								
	4. Place on table			&  5&			 &x$^1$					&			&x$^1$			&x 			&	x				&						&				& 				&   		&  			  &  			&  					&				& 				&				& 	x			&	x		& 					&					&				& 					&				&				&						 &				    & 						&				&				&					&			\\ \hline
	5. (a) Flatten cloth		&  2&			 &						&			&				&			&	x				&						&				& 				& 			&  			  &  			&  					&				& 				&				& 				&			& 		x$^4$		&					&				&                   &				&				&						 &				    & 						&				&				&					&			\\ \hline	\rowcolor{LightGreen}																																																																																																																																								
	5. (b) Flatten cloth		&  1&			 &x$^1$					&			&				&			&					&						&				& 				&   		&  			  &  			&  					&				& 				&				& 				&			& 					&					&				&                   &				&				&						 &				    & 						&				&				&					&			\\ \hline
	5. (c) Flatten cloth		&  2&			 &						&			&				& 			&					&						&				& 				& 	 		&  			  &  			&  					&				& 				&				& 				&			& 					&	x				&		x		&                   &				&				&						 &				    & 						&				&				&					&			\\ \hline	\rowcolor{LightGreen}																																																																																																																																								
	5. (d) Flatten cloth		&  1&			 &						&			&				& 			&					&						&				&				&   		&  			  &  			&  					&				& 				&				& 	 			&			& 					&					&				&                   &				&				&						 &				    & 			x			&				&				&					&			\\ \hline
	6. Separate from pile		&  3&			 &						&			&				& 			&					&						&				& 				&   		&  			  &  			&  					&				& 				&				& 				&			& 					&					&				&      x            &	x			&	x			&						 &				    & 						&				&				&					&			\\ \hline	\rowcolor{LightGreen}																																																																																																																																								
	7. Pick \& place fd cl 		&  2&			 &x$^1$					&			&				& 			&					&						&				& 				&   		&  			  &  			&  					&				& 				&				& 				&			& 					&					&				&                   &				&				&	x$^5$				 &					& 						&				&				&					&			\\ \hline
	8. Clamp (p)flat cl			&  5&			 &						&			&				&			&	x				&						&				& 	x$^2$		&   		&  			  &  			&  					&				& 				&				& 				&			& 					&					&				&                   &				&				&						 &	x				& 			x			&		x		&				&					& 			\\ \hline	\rowcolor{LightGreen}																																																																																																																																								
	9. Pinch (p)flat Clamp		&  3&			 &						&			&				&			&					&						&				& 				& 	 		&  			  &  			& 					&				&				&				& 				& 			& 					&					& 				&                   &				&				&						 &	x			    & 			x			&		x		&				&					&			\\ \hline
	10. Pick single layer		&  1&			 &						&			&				& 			&					&						&				& 				& 	 		&  			  & 			&  					&				& 				&				& 				& 			& 					&					&				&                  	&				&				&						 &				    & 						&		x		&	 			&					&			\\ \hline	\rowcolor{LightGreen}																																																																																																																																								
	11. Grasp gaiting			&  1&			 &						&			&				& 			&					&						&	x			& 				&   		&  			  &  			&  					&				& 				&				& 				&			& 					&					&				&                   &				&				&						 &				    & 						&				&				&					&			\\ \hline
	12. Hang 					&  1&			 &						&			&				&			&					&						&				& 				&	 		&  			  &  			& 					&				&				&				& 				& 			& 					&					&				&                   &				&				&						 &				    & 						&				&		x		&					&			\\ \hline	\rowcolor{LightGreen}																																																																																																																																								
	13. Put hanger 			 	&  1&			 &						&			&				& 			&					&						&				& 				&   		&  			  &  			&  					&				& 				&				& 				&			& 					&					&				&                   &				&				&						 &					& 						&				&				&		x			&			\\ \hline 
	14.Japanese fold 			&  1&			 &						&			&				& 			&					&						&				& 				&   		&  			  &  			&  					&				& 				&				& 				&			& 					&					&				&                   &				&				&						 &					& 						&				&				&					&	x		\\ \hline	\rowcolor{LightGreen}																																																																																																																																								
	TT    						&   &	1		 &	5					&	1		&	3			& 	3		&	4				&		1				&	2			& 	2			& 	2 		&  	1		  & 	1		&  1				&	1			& 1				&		1		& 		1		& 	1		& 			1		&		2			&		1		&       1          	&	1			&		1		&		1				 &		2		    & 			3			&		3		&	 	1		&			1		&	1		\\ \hline	
																																												
\end{tabular}
\newline

$1.$ Open loop procedure.
$2.$ Manipulation to first bend the edge to be able to clamp.
$3.$ Applying controlled tension between grasps.
$4.$ Single handed (without H2 holding).
$5.$ Single handed.

\end{table*}

If we consider the extrinsic geometries they use, the grippers in \autoref{fig:grippers}.\textbf{b} and \textbf{v} can perform a \PiPie{} grasp, the gripper in  \autoref{fig:grippers}.\textbf{i} a \LPie{}  and \PPie{} grasps and the gripper in \autoref{fig:grippers}.\textbf{u, l} and \textbf{v} a \PPPie{} grasp, necessary to perform a pinch of a cloth that is lying flat on a table. The gripper \autoref{fig:grippers}.\textbf{u} is specially designed to be able to perform simultaneously a \PP{} and a \PPPie{} grasp, used to grasp a folded corner while holding the cloth under. 
The most versatile gripper in terms of possible textile grasps is the one in \autoref{fig:grippers}.\textbf{v} from ~\cite{ono2007better} that can perform the grasps \PiPie{},  \PPi{}, \PP{} and \PPPie{}. \autoref{tab:graspsInPapers} summarizes the relationship between all the reviewed works, what gripper they use and how many grasps they can realize.

\autoref{fig:taxonomy} shows all these grasps represented in a schematic way, as well as the grasp combinations appearing in the reviewed bibliography. Note that bimanual robots performing cloth manipulation tasks in the literature have dealt  with up to 5 virtual fingers, leading to up to three couples of opposition forces.

\subsection{Encoding previously tackled tasks within the framework} \label{subsec:applicationOnTasks}

 In \autoref{tab:tasks} we list all the tasks that we have identified in  literature related to cloth folding. For each task, we define the corresponding sequence of primitives and label each primitive with its type, using the nomenclature proposed in \autoref{sec:framework}. In addition, \autoref{tab:tasksInPapers} shows what works address which tasks.

Unfold in the air (task 1(a)) consist in grasping a crumpled cloth and regrasp it until its gasped flat by two significant points. It has been solved in ~\cite{osawa2007unfolding,maitin2010cloth,kita2011clothes,bersch2011bimanual,li2015regrasping,Doumanoglou2016}. These are all vision-based solutions where grasping points have to be located to regrasp the cloth by the corners (or other grasping points) to permit its recognition. This task relies on vision and recognition algorithms, more than on the manipulation itself. A good review of such algorithms can be found in ~\cite{sanchez2018robotic}. As regards to manipulation, what is important is the gripper ability to grasp the correct point without changing the configuration of the cloth during the approach. In a similar way (task 1(b)), some authors ease the recognition and grasping of cloth corners by sliding the cloth on a table ~\cite{cusumano2011bringing}, leading to a slightly modified strategy of the same task (task 1(b)). The same objective of tasks 1(a) and (b) is accomplished in task 1(c) simplifying the vision requirements and increasing the challenge of manipulation. It is based on localizing only one corner on a table (provided it is visible), grasp it, and reach for another corner with the second hand by sliding the gripper along the edge of the cloth without losing it ~\cite{yuba2017unfolding}. To increase the chances of having at least one corner visible, the authors previously slide the cloth on the table, similarly as in task 1(b). This edge tracing manipulation can be faster because it reduces the number of grasping points that need to be localized, but the manipulation of "pinching and sliding" is of high complexity, in many cases requiring specially designed grippers ~\cite{sahari2010Clothes,shibata2012fabric}. Unfolding in the air is a crucial task to recognize the manipulated cloth and set the canonical cloth state to start any other manipulation, therefore, it is one of the most popular tasks, appearing 10 times in \autoref{tab:tasksInPapers}.

Once the cloth is unfolded in the air, placing it on the table (task 4) can be done with an open-loop procedure, but challenges may arise at planning time if relying on a cloth model that needs to accurately deal with friction and collisions ~\cite{ruan2018accounting}. Moreover the task may require taking into account complex dynamics to ensure the cloth is laid flat ~\cite{balaguer2011combining}.

 Only three of the tasks involve folding itself: fold on table, fold in the air and dynamic fold in the air (tasks 2, 3(a) and 3(b)). The first has been the second most attempted task in literature. Fold on table consists in grasping two vertices and folding the cloth until the desired fold is attained ~\cite{Doumanoglou2016, li2015folding, miller2012geometric}. Planning such folds requires recognizing the cloth and deciding where the folding lines should go ~\cite{miller2012geometric} but also to compute the hand trajectories to realize the manipulation ~\cite{li2015folding}. A big linear gripper can simplify the manipulation ~\cite{Foldy2010} but is not general enough to deal with several pieces of cloth of different sizes.  Another challenge is to monitor how the cloth is being folded in a controlled manner, for instance, with vision ~\cite{jia2017manipulating} or with force control manipulation maintaining tension between hands ~\cite{lee2015learning}.
The task of folding in the air is tackled in ~\cite{colome2018dimensionality} and ~\cite{bersch2011bimanual}. It presents several manipulation challenges that are not fully solved in the papers; for instance, the grasping and releasing is not addressed in ~\cite{colome2018dimensionality} and the task is done as an open-loop procedure in ~\cite{bersch2011bimanual} without assessing if the T-shirt is properly folded. This approach to folding promises to be much faster than the other, but it may require different procedures depending on the cloth item. The dynamic fold done in ~\cite{yamakawa2011motion} is also fast, but entails a very complex and risky manipulation, and it requires a high-velocity hardware.

Cloth flattening on the table has been attempted by 7 authors, making it the third most popular task, which has been tackled with several different approaches (task 5(a to d)). Some works slide surfaces of the grippers on top of the cloth ~\cite{maitin2010cloth, Doumanoglou2016,cuen2008action} and others grasp edges or vertices and pull them  ~\cite{willimon2011model,sun2015accurate}. The challenges are to understand the state of the cloth, localize the creases and define a direction of sliding or pulling to flatten the cloth. A more challenging task from the manipulation point of view would be that of unfolding a corner that has remained folded. Such grasp is difficult because it is easy to grasp at the same time the cloth that lies under the corner. A specially designed gripper for this task is presented in ~\cite{koustoumpardis2014underactuated} (\autoref{fig:grippers}.\textbf{u}) which uses one finger to hold the cloth underneath and another finger to grasp the corner on top.

Another important task is pick and place. Regarding crumpled clothes, pick and place becomes challenging when only one cloth needs to be picked up from a bunch of several tangled clothes (task 6). Again, the main challenge in this task comes from vision ~\cite{ramisa2012using, hamajima2000planning}. Picking up folded clothes, a task that would be very relevant in any domestic environment, was done in ~\cite{maitin2010cloth} using an open-loop procedure and it is approached in more detail in ~\cite{moriya2018method}, but only using one hand. The paper focuses on localizing the edge of the folded piece that doesn't have loose layers and, therefore, it can be picked without being unfolded. 

Other popular tasks are those of clamping and pinching (tasks 8 and 9). A pinch consists in grasping the cloth from one side, while in a clamp each finger is in a different side of the cloth. The main challenge for the clamp is to go under the cloth before grasping while, for a pinch,  a \PPPie{} grasp needs to be performed and then the cloth dragged by bringing the two point contacts together. The same strategy for pinching could also lead to a clamp ~\cite{koustoumpardis2014underactuated}.  Works like ~\cite{shibata2012fabric}  create first a crease in the cloth to facilitate clamping. Note that a pinch or a clamp is necessary to grasp the cloth to fold it on a table (task 2), but not all the reviewed works addressing task 2  appear in \autoref{tab:tasksInPapers} for clamping or pinching because some of them don't care if the grasp was a clamp or a pinch as long as it was close to the edge ~\cite{maitin2010cloth,miller2012geometric,li2015folding}. However, this distinction may be more relevant when the quality of the end result becomes more important. Nowadays, it is taken into account in papers whose main focus is to design a gripper for textile manipulation ~\cite{koustoumpardis2014underactuated, kabaya1998service,Le2013development }. 

The last five tasks in the list appear only in one paper. Picking a single layer is approached in ~\cite{ono2001picking} by pressing on the pile to create more spacing allowing to clamp one single layer. However, this approach has been solved in  industry with vacuum grippers quite efficiently. Grasp gaiting could substitute the "pinch and slide" to trace an edge, so as to reduce the risk of losing the cloth ~\cite{sahari2010edge}. However, it will probably turn out to be too slow compared to sliding to constitute a practical solution. 

Finally, the tasks of hanging clothes or putting hangers inside garments are of obvious utility in domestic and commercial environments and they will probably attract more attention in the near future.


\begin{table*}[th]
    \caption{Describing manipulation primitives in \autoref{tab:tasks}}
    \label{tab:primitives}
    \centering
\setlength\tabcolsep{3pt} 
\renewcommand{\arraystretch}{1.15} 
\begin{tabular}{|l|r|}
\hline
\textbf{Primitive} & \textbf{\#} \\
\hline
\multicolumn{2}{|c|}{\Ex Adding/removing extrinsic contact} \\
\hline
Add \Pe{} to  2\PP{} 	& 1\\
Add \Pie{} to 2\LL{}, 2\PP{} or \PP{} & 1,2,2	\\
Rm \Pie{} from 2\PP{}+\Pie{} &	2 \\
Rm \Pie{} from \PP{}+\Pie{} &	2 \\
& \\
\hline
Total & 10\\
\hline
 \hline
\multicolumn{2}{|c|}{\G Grasping} \\
\hline
Gr \PPie{} or sh 2\PPie{}	& 1,2\\
Gr 2\PPie{} or 2\PiPie{} & 1,1 \\
Gr \PiPi{}      & 1\\
Gr 2\PP{}	            & 1\\
Gr 2\PP{} + \Pie{}      & 2\\
Gr \PP{} + \Pie{}       & 4\\
&\\
\hline
Total                   & 13\\
\hline
\end{tabular}
\begin{tabular}{|l|r|}
\hline
\textbf{Primitive} & \textbf{\#} \\
\hline
\multicolumn{2}{|c|}{\R Releasing} \\
\hline
 Release \PP{}, 2\PP{} or \PiPi{}                    & 1,1,1 \\
 Release 2\LL{} from inside a fold  & 1	\\
 Release \PP{} or 2\PP{}  on a table             & 2, 4\\
 Release \PPie{} or  sh 2\PPie{} 	                    & 1,1\\
& \\
 \hline
 Total                              & 12\\
\hline
 \hline
\multicolumn{2}{|c|}{\RG Regrasping} \\
\hline
 From 2\PP{} to \PP{} and viceversa	                & 2,4 \\
From \PP+\Pie to 2\PP{}+\Pie            & 1\\ 
From \PP{} to \PP{}+\Lint{}               & 1	\\
 From \PP{}: Add point to \PP   	    & 2\\
 From sh 2\PPie{} to  \PP{}+sh 2\PPie	& 1\\
 From sh 2\PPie{} to \PP{} + \Pie{}     & 1\\
 & \\
 \hline
 Total &                                12 \\
 \hline 
 \end{tabular}
\begin{tabular}{|l|r|}
\hline
\textbf{Primitive} & \textbf{\#} \\
\hline
\multicolumn{2}{|c|}{\inGM In grasping motions} \\
\hline
 in 2\PP{}+\Pie{}: H1 folds \& H2 holds or both Hs fold & 1,1 \\
 in \PP{}+\Pie{}: Pull to flatten or unfold the gr. corner & 1,1 \\
 in 2\PP{}: Shake to move hanging v. & 2 \\
 in 2\LL{}: Pronation rotation of each H & 1 \\
 in \PPie{}: Push inwards cloth (to create crease) & 1 \\
 in sh 2\PPie{}: Bring P contacts together & 1 \\
 in \PP{}+\L{}: Apply tension between hands & 1 \\
 in \Pie{}: Move gripper finger under cloth & 1 \\
 \hline
 Total & 11\\
  \hline 
   \hline
\multicolumn{2}{|c|}{\Sl Sliding} \\
\hline
in 2\PiPie{}: Slide H1 \& H2 in opp. dir.    & 1 \\
in 2\LPie{} or 2\PPie{}: Slide H1, hold H2  & 1 \\
in 2\PP{}: Trace edge                       & 1 \\
in \PP{}+\Pie: Slide cloth on table to reveal corner       & 1 \\
 \hline
 Total & 4\\
 \hline
\end{tabular}

\end{table*}

\begin{figure}[tb]
    \centering
   \includegraphics[width=\linewidth]{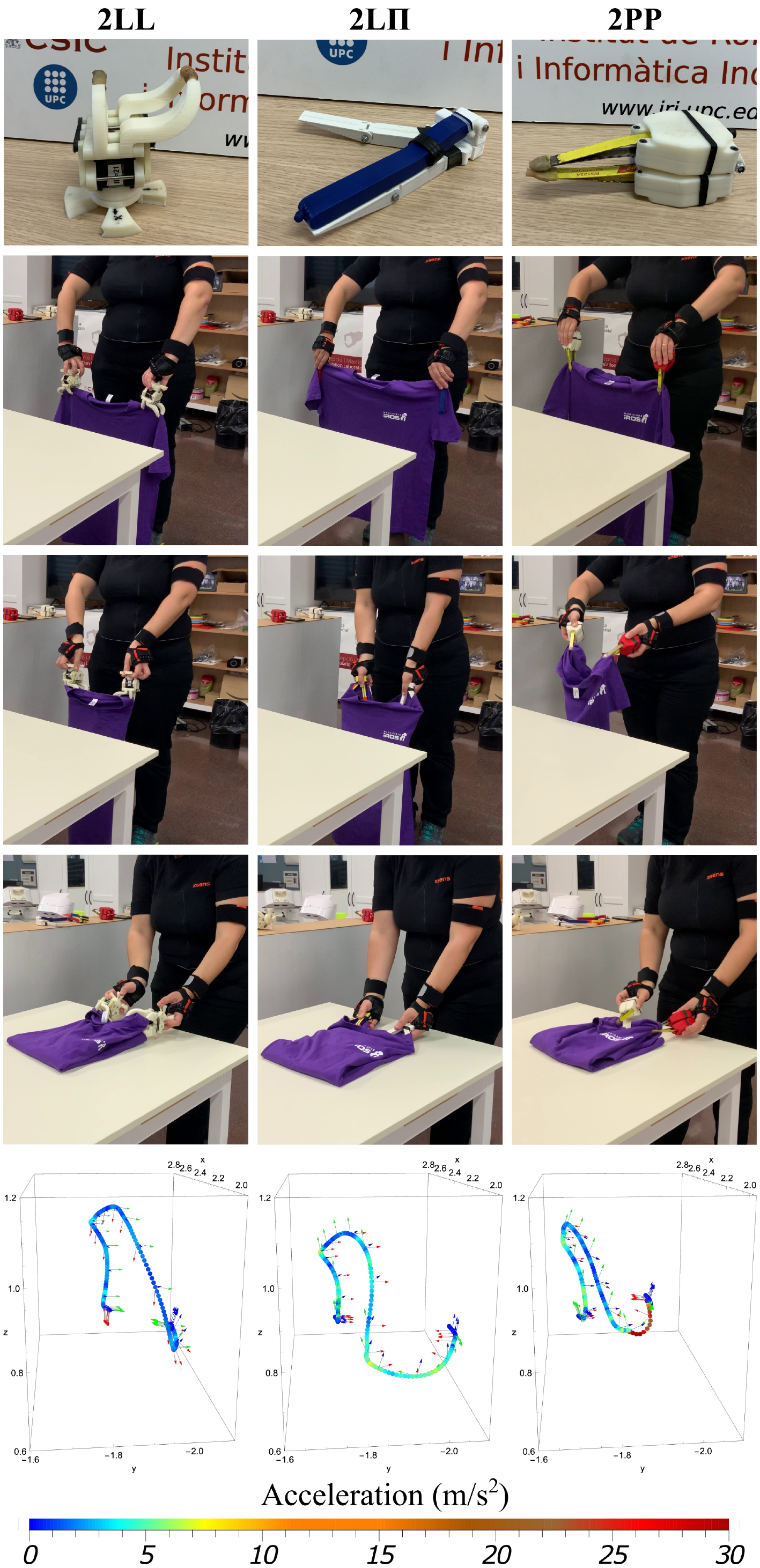}
    \caption{Setup of the experiments with three different grasp types to fold a T-shirt in the air. The left column shows frames of the task using the \LL grasp, the middle column for the \LPi grasp and the right one for the \PP grasp. Bottom line shows the trajectories corresponding to the right gripper of one trial, with color code representing acceleration norm. }
    \label{fig:experimentSetup}
\end{figure}

\section{Applications of the framework} \label{sec:applications}

\subsection{Study of manipulation parameters under different grasp types when folding a T-shirt in the air}\label{subsec:experiment}

So far, we assume and claim that choosing one type of grasp or another among those we have defined could greatly influence the performance of a task. To study this influence in more detail we have designed  a simple experiment where we recorded motion data of a subject \textit{wearing} different types of grippers, using the XSens suit and performing a cloth manipulation task. The hands were rigidly holding the grippers, preventing relative motion between hand and gripper.

We compared the three grippers that are shown at the top of \autoref{fig:experimentSetup}. The first is the evolution of the gripper that appears in \autoref{fig:grippers}-o, it is a parallel gripper where each closing part consists of two points. Assuming two points form a line, we can define that grasp as a \LL{}. The next to the right gripper consists of 2 rigid lines at the bottom, forming a plane, and a linear finger on top. Therefore, that gripper can perform a \LPi{} grasp. Finally, the last gripper performs a simple \PP{} grasp. All the grippers are used passively with a human-in-the-loop manipulation, and we assume the cloth is pre-grasped. The task performed is described as the task 3 in \autoref{tab:tasks}, "Fold a T-shirt in the air", but in this case there is no release step.

\begin{figure}
\centering
       \includegraphics[width=\linewidth]{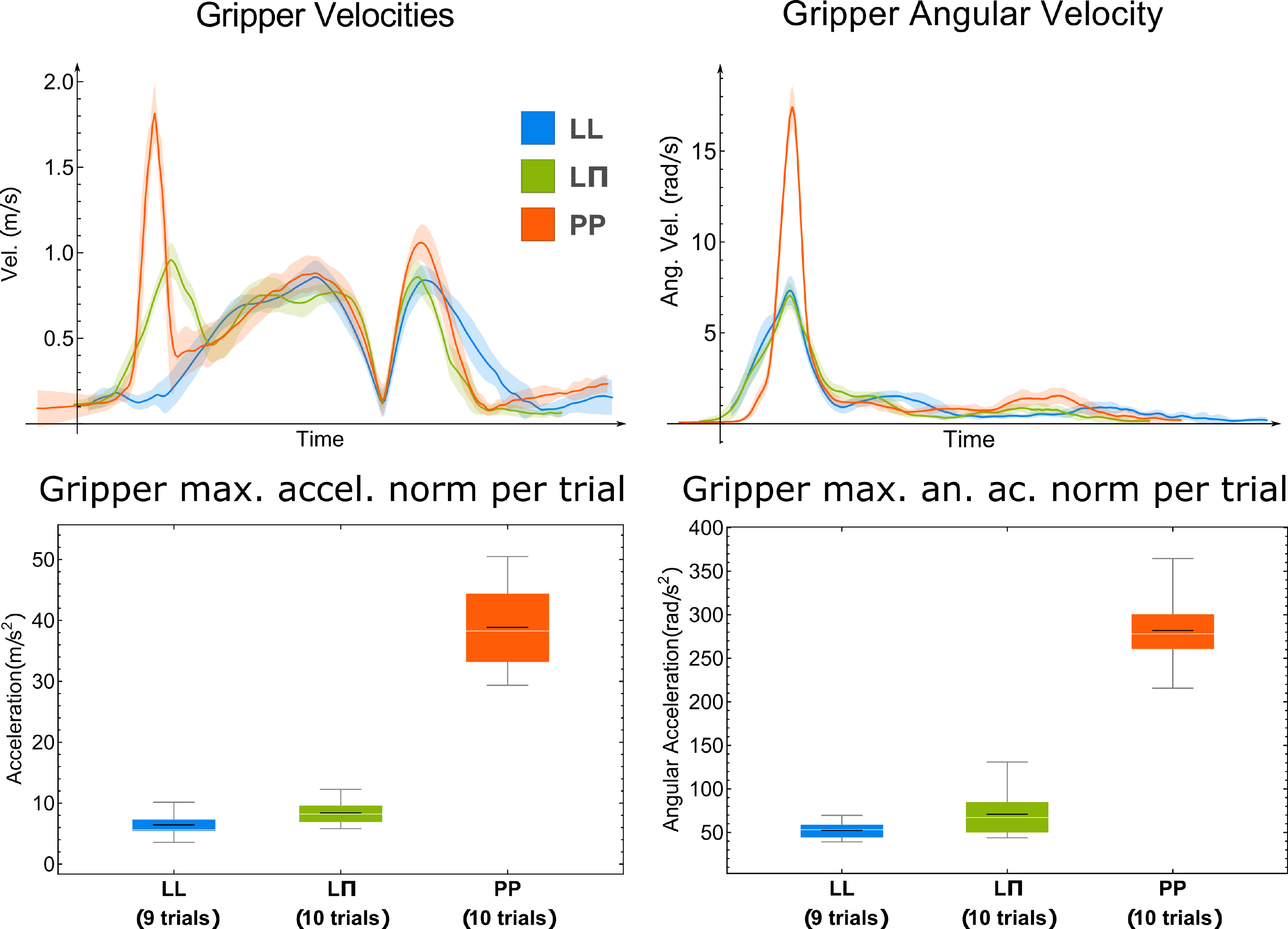}
    \caption{(Top) Comparison between the norms of the linear velocity (left) and the angular velocity (right) of the grippers during the executions of the trials. (Bottom) Comparison between peak accelerations (left) and angular acceleration (right) of the grippers during execution of the fold in the air.}
    \label{fig:results}
\end{figure}

We repeated the tasks 10 times, wearing the three grippers.  We collected upper body motion data with the Xsens suit, using an extra sensor attached to the right gripper to directly record its position, velocity and acceleration. This extra sensor is part of the Xsens suit and is usually used to record motion of a manipulated object or a tool, like for instance, a sword. The bottom graphics in \autoref{fig:experimentSetup} show example trajectories corresponding to one trial of each gripper, each point colored with the corresponding linear acceleration norm. In
\autoref{fig:results}-(top) we show the different velocity profiles for each gripper. Each curve and shaded area represents the mean and standard deviation of all the trial curves, aligned to match the maximum and minimum peaks. At the bottom, we compare the maximum acceleration peaks for the trials performed with the different grippers.

The results of the experiment show evidence on how grasp type can influence the execution of a task. When the gripper provides more support surface along the T-shirt shoulders, the task of folding requires less acceleration, indicating that the dynamics of the task is greatly simplified. We cannot observe a significant difference between the \LL{} gripper and the \LPi{} gripper, they both provide sufficient support and therefore, the acceleration peak is similar. Instead, we can appreciate differences in the trajectories (bottom of \autoref{fig:experimentSetup}) because the rotation pivot point is aligned with the rotation axis of the wrist for the \LL{} gripper, while with the other gripper the user tends to rotate it about the abducted fingertip.

 Finally, at the last frames of the task shown in \autoref{fig:experimentSetup}, we can observe that the quality of the fold is also different depending on the gripper used, the results being clearly worst for the \PP{} grasp. We could observe this, but we haven't included the results because we do not have a fair measure to compare qualities of fold.

\subsection{Increasing versatility and task repertoire} \label{subsec:discussion}

With the experiment in the previous section we have seen how the dynamics of the task of "Fold a T-shirt in the air" can be greatly simplified by just using the appropriate grasp type. This can reduce the training times when applying learning techniques, and also change the quality of the result. But there are more examples where
improving manipulation skills can reduce the complexity of many of the perception and pose estimation algorithms.
For instance, the task "Unfold in the air" 
may seem relatively easy by looking at the manipulation primitives used, but finding and localizing the two corners and successfully grasping them is a difficult endeavor  that usually involves complex vision and recognition algorithms. 
A way of reducing the complexity of recognizing the second grasp corner is for the second hand to grasp an edge point close to the first grasp point, and then trace the edge until the next corner is reached, which can potentially be faster than looking for the second corner. This is at the expense of gaining manipulation complexity, meaning that more attention has to be put on touch sensors and increasing contact surfaces to gain more tactile feedback.


There are many other  examples of the importance of versatile grasping abilities. The task "Fold on table", for instance, has been always addressed in a very similar way: grasping two corners laying on the table and performing a coordinated motion with the arms. However, this may not be the best approach and sometimes it doesn't yield very good end results. Other approaches could consist in folding with one hand while the other holds the cloth by the folding line. However, for that to happen we probably need a linear grasp on the edge and a planar or linear hold along the folding line, which is not possible with most of the off-the-shelf existing grippers. 

Moreover, this approach to folding can't be applied to large clothes like bed sheets or tablecloths. We  humans usually fold them in the air, starting from an unfolded 2\PP{} grasp, to bring the two corners together, regrasp them in a single hand and use the other hand to trace the edge until the end of the fold, to end up again in a 2\PP{} grasp where the cloth has one extra fold. When several of these folds have been done, the last one to finish can be done on the table. For big bed sheets, this same strategy is started with 4 hands (2 people). This is of course a complex manipulation that  requires more versatile grippers to be executed successfully and more manipulation skills. However, this kind of operations, or something equivalent, will be needed eventually if robots are to fold big textile objects.


Alternative grasps could also be needed depending on the cloth material.
As an example, the task of picking and placing a folded garment (task 7 in \autoref{tab:tasks}) has been implemented in ~\cite{maitin2010cloth} using a 2\PP{} grasp. This was successful for the folded towels used in the paper, but they required to be flattened after being placed on the table. We believe that attempting this task with only point contacts would not work for a thinner textile material, as it would unfold while transferring it. Using a planar contact under the garment can provide better support to achieve the task, for instance a 2\PPi{} or 2\LPi{} grasp, and prevent it from unfolding while transferring it or having to flatten it after begin placed. In ~\cite{moriya2018method} the task is performed with a single \PP{} grasp using a parallel gripper. Provided the correct edge is grasped, one can lift the folded cloth without unfolding it, but to place it back one needs to perform a motion similar to "Place flat on table" (task 4) to prevent unfolding the cloth. This motion requires free space to allow the arm to move. However, if the task requires to place the folded clothes inside a shelf, as it may be expected for folded clothes, that space for the motion may not be available. Therefore, here using a  2\PPi{} or 2\LPi{} grasp becomes again relevant to reduce the workspace volume needed to perform a task.

Extrinsic contacts are fundamental in cloth folding, for instance to perform a pinch. Other strategies have been explored for rigid objects in ~\cite{Eppner2015} and they should be explored and evaluated in the context of cloth manipulation. For instance,  bringing the object to the table edge to facilitate clamping. This  has only been considered in ~\cite{koustoumpardis2014underactuated} as a scenario to evaluate a novel gripper for clothes, but should receive more attention to evaluate its usefulness with respect to other strategies.

In conclusion, we believe that to achieve a more varied repertoire of manipulations, increase success ratios and improve the quality of results, we need, among other things, to be capable of executing more diverse and sophisticated grasps.

\subsection{Providing guidance for gripper design} \label{subsec:gripperDesign}
The definition of grasps for textiles using geometric virtual fingers permits expanding the taxonomy in \autoref{fig:taxonomy} with as many combinations of geometries one can think of. In the previous section we have already identified some tasks that could use other geometries, like "Pick \& place folded cloth" could use a plane to support the safe transfer of the object or "Fold a T-shirt in the air" benefits from a linear geometry that can transmit more torque to bend in the shoulders of the T-shirt.

\autoref{tab:primitives} collects all the manipulation primitives appearing as task steps in \autoref{tab:tasks}, which add up to 62 primitives. The table classifies them by type and counts the multiplicity of each one. We can see how most of the primitives appear only once, indicating the complexity and particularity of each manipulation task. This may explain the lack of general solutions that are found in cloth manipulations, as pointed out in ~\cite{sanchez2018robotic}.

It is striking that the implementation of manipulation primitives has so strongly relied on the PP grasp so far. This is probably due to the fact that generic parallel-jaw grippers are the most available and, when designing grippers to handle textiles, they were tailored to a very specific application. More versatile grippers able to deal with cloth could achieve many more primitives with different and more tailored functionalities to ease reaching the objective of every task.

Identifying all the individual primitives that we want a gripper to execute can help to identify the requirements each grasp will need in terms of friction, force or precision, which are to be taken into account when we design grippers. For instance, sliding on cloth or releasing from inside a fold require low friction, but hanging a cloth with a \PP{} grasp may require a lot of force, or increased friction. Grasping a cloth corner requires precision while a 2\PPi{} to pick up a folded cloth may not require to grasp a particular point. 

In our work ~\cite{donaire2020versatile,donaire2019ma} we defined gripper requirements using our framework, selecting three tasks and extracting their required grasp geometries. We selected to implement a \LL{} geometry with one linear finger on top, and two bottom finger that can abduct to morph the gripper into a \LPi{} geometry, similar to the gripper in the top middle of \autoref{fig:experimentSetup}. We aimed at providing the skills for performing three tasks: picking and placing folded clothes using a \LPi{} grasp, the edge tracing task and folding a T-shirt in the air with this same \LPi{} grasp. To execute all the primitives we needed variable friction capacities, so we took inspiration from the design in ~\cite{spiers2018variable}. Our design method consisted in implementing progressive simple motor-less prototypes to test the geometries, like the one shown in the top-middle picture in \autoref{fig:experimentSetup}, and to better define the requirements for each primitive. Finally, we implemented a fully functional prototype that could execute the required tasks.

Aiming at versatility in the design of grippers to handle textiles is a very challenging goal which, up to the authors' knowledge has been addressed only by very few authors before and was solved only partially ~\cite{ono2007better, koustoumpardis2014underactuated}, probably because there hasn't been a deep study of the grasping and manipulation variety needed to manipulate clothes. Our framework offers a systematic way to explore grasp combinations, with potential to identify the most useful to enhance manipulation primitives, thus suggesting versatile end-effector designs, including multigrip and reconfigurable grippers able to reshape themselves into virtual fingers with different geometric contacts.

\subsection{Identifying and characterizing manipulation primitives}

Following \autoref{subsec:grasping_manipulation},  a given task can be segmented according to the used grasps, to obtain different segments that correspond to manipulation primitives like the ones in \autoref{tab:primitives}. Combined with motion capture approaches, this can provide large amounts of labeled manipulation data, provided we can either recognize grasp type from the data (using data gloves) or by manually labeling them.


Assigning semantic tags for each of the primitives, as it is done in \autoref{tab:primitives}, allows to learn semantic sequencing of steps within tasks, with applications to autonomous high-level task planning. On the lower level,
having data about trajectories, velocities and accelerations for each of the primitives enables to learn motion representations, for instance DMPs, for each of them. Therefore, this permits defining motion building blocks that can be  sequenced to perform a given task. Similar approaches have been successfully applied in literature for rigid object manipulation ~\cite{aksoy2011learning,aein2018library} or whole-body human motions ~\cite{kulic2012incremental}.

Our framework opens the door to apply these kind of methods to cloth manipulation tasks.  In order to successfully implement this kind of approaches fully from collecting data to the execution level, several additional problems need to be solved, including efficient action representation for each of the primitives and efficient cloth state representation, perception and estimation. In our research group we are working in parallel on several of these problems in the context of the ERC Clothilde project.

\subsection{Benchmarking manipulation}
One of the great challenges of benchmarking manipulation is to understand the complexities of all the possible skills that need to be evaluated, thus classification efforts are needed to set up the grounds for a proper benchmark. Several examples exist in literature where classifications like grasping taxonomies ~\cite{Feix2016Grasp}, ontologies of actions ~\cite{worgotter2013simple}, or  task taxonomies ~\cite{quispe2018taxonomy} are used to evaluate robot performance in carrying out tasks, mostly involving rigid objects. 

We are working on using the proposed framework to classify and evaluate the complexity of a given task. Assessing the complexity of a task is useful to rank tasks in a benchmark and to enable  their evaluation, specially if the evaluation has to be done qualitatively. In other words, if we can just state if a task was successfully completed or not, a quantification of the complexity of the task can attribute value to a solution.

In addition, our framework puts special effort in abstracting from the human hand shape and the robot embodiment, which is of special importance when benchmarking manipulation, as many research labs have different robotic systems solving similar problems, and comparison of results must be independent of the used platform.


\section{Conclusions and future research} \label{sec:conclusions}

We have proposed a framework to characterize and systematize grasps, manipulation primitives and tasks for the versatile handling of clothes by robots. Grasps have been redefined in terms of prehension geometries, that we called geometric virtual fingers. This generic definition abstracts from the particular robotic embodiment that is used and can easily inspire gripper designs. Because cloth will conform to the shape of the grasping geometry, this grasp definition is not influenced by object shape or size. Instead, our definition focuses on the geometric shape of the gripper contact area where the robot has to transmit forces to the hanging part of the cloth. In the specialized literature only some combinations of geometries have appeared, the \PP{} being the most prominent one by far. Similarly, the most used grippers can only implement a basic pinch.

We have identified several tasks that are solved in previous works using pinch grasps, which could be improved if more versatile grippers were used. The following types of improvements are envisaged: 

 \begin{itemize}
 	\item {\it More efficient learning and execution of a given task.} For example, when folding a T-shirt in the air, an L contact geometry could help bend each shoulder, thus transmitting the appropriate dynamics to the hanging part of the shirt, and reducing required dynamics of the robot arm.
 	\item {\it Higher quality of the end result.} A clothing item could be better folded if a folding line were rigidly maintained, either intrinsically or extrinsically, or tension were applied between parts of the cloth. 
 	\item {\it Wider range of textile materials for which a task could be completed.} Providing larger support surfaces or multiple well-placed contacts could ensure a suitable end result for objects of varied fabrics.
 	\item {\it Increased repertoire of robotized tasks.} For instance, it is not possible to fold a big piece of cloth (e.g., a bed sheet) or turn inside out a shirt with state-of-the-art procedures.
 	\item {\it Reduced workspace requirements.} Clothes could be placed on a tight shelf or a drawer if a gripper implementing 2\PPi{} or 2\LPi{}  grasps were available, capable of compressing a pile of folded clothes.
 	\item {\it Reduced auxiliary technical requirements.} Dependency on complex vision methods could be reduced by relying on robust manipulation primitives such as sliding to trace edges or flatten a cloth lying on a table.
 \end{itemize}
 
 In addition, our framework considers both intrinsic and extrinsic geometries, allowing for the definition of extrinsic dexterous manipulations in a natural way, just as additional re-grasping operations, that is, particular instances of manipulation primitives as appearing in Table IV.

%
%
Our analysis has shown that many relevant cloth manipulation tasks have been only barely addressed so far or require major improvements. 
Identifying the manipulation primitives and the grasps required by each task following the presented framework will help determine the kind of gripper and the actions a robot needs to be able to perform. Moreover, such systematic task specification is a step towards establishing a benchmark for cloth manipulation, in which alternative strategies could be explored and evaluated in terms of success ratios and quality of end results.

We are currently carrying out a human study --using a motion capture system and some hand-held grippers-- to evaluate what grasps humans use to execute a range of cloth handling tasks. Such grasps will then be encoded in terms of geometric virtual fingers and the study will serve to validate and refine our framework, by perhaps expanding the list of grasps and tasks, so as to ultimately generate a complete benchmark of cloth manipulation primitives. 



\bibliographystyle{ieeetr}

\begin{IEEEbiography}[{\includegraphics[width=1in,height=1.25in,clip,
keepaspectratio]{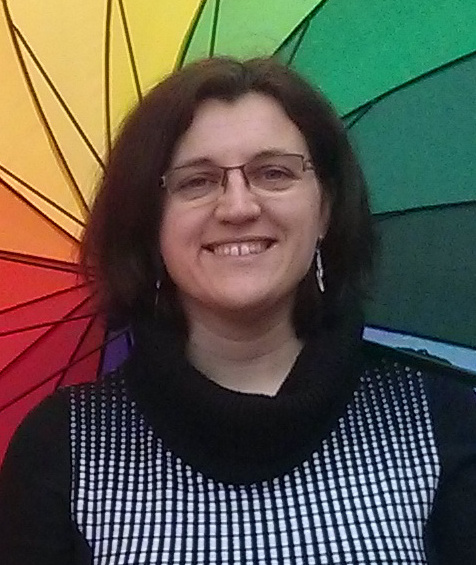}}]{J\'ulia Borr\`as} is a  Mathematician and Computer Scientist since 2004 and 2006, respectively, and she obtained her European Ph.D. degree in 2011 working on kinematics and reconfiguration designs for the Stewart-Gough parallel platform. She worked abroad for 6 years as a postdoc, two years at Prof. Aaron Dollar GrabLab group from Yale University and four years at the Karlsruhe Institute of Technology (KIT) with prof. Tamim Asfour H2T group. She has worked on parallel robots, underactuated robot hands, grasping, dextrous manipulation, whole-body motion analysis, humanoid robot locomotion, novel designs for robotic hands and grippers, and robotic cloth manipulation.  In 2018 she was awarded a Ramon y Cajal scholarship, one of the most prestigious senior postdoctoral scholarships in Spain. Recently, she has earned a tenured position at the Spanish Scientific Research Council (CSIC). 
\end{IEEEbiography}

\begin{IEEEbiography}[{\includegraphics[width=1in,height=1.25in,clip,
keepaspectratio]{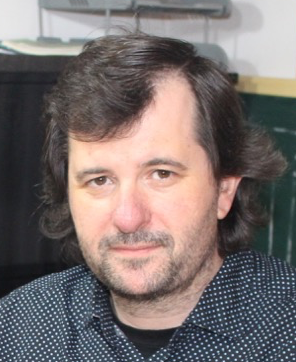}}]{Guillem Aleny\`a} holds a tenured position at Spanish Scientific Research Council CSIC. He received a PhD degree (Doctor Europeus) from UPC in 2007 with a work on mobile robot navigation using active contours while he was supported by a EU-FP6 Marie-Curie scholarship. He has been visitor at KIT-Karlsruhe, INRIA-Grenoble and BRL-Bristol. He has coordinated numerous scientific and technological transfer projects involving image understanding, task learning and plan execution tasks. He has published more than 100 articles in relevant venues in the areas of robotics, computer vision, and artificial intelligence.
\end{IEEEbiography}

\begin{IEEEbiography}[{\includegraphics[width=1in,height=1.25in,clip,
keepaspectratio]{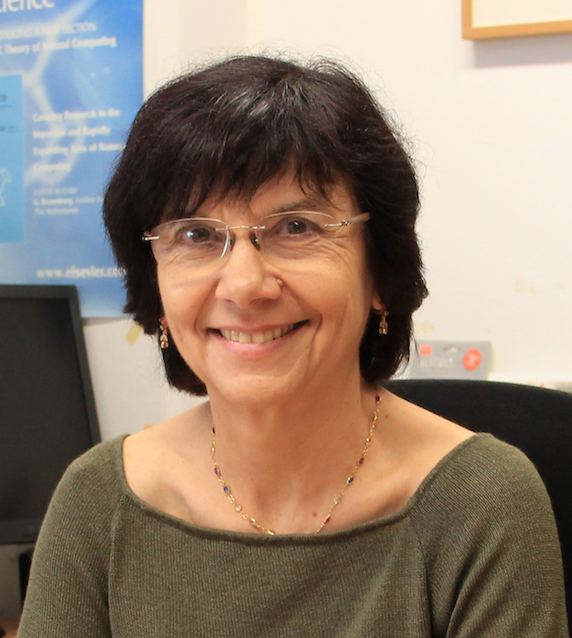}}]{Carme Torras} is Research Professor at the Institut de Robòtica i Informàtica Industrial (CSIC-UPC) in Barcelona, where she leads a research group on assistive and collaborative robotics. She received M.Sc. degrees in Mathematics and Computer Science from the University of Barcelona and the University of Massachusetts, respectively, and a Ph.D. degree in Computer Science from the Technical University of Catalonia (UPC). Prof. Torras has published six research books and more than three hundred papers in robotics, machine learning, geometric reasoning, and neurocomputing. She has supervised 19 PhD theses and led 16 European projects, the latest being her ERC Advanced Grant project CLOTHILDE – Cloth manipulation learning from demonstrations. Prof. Torras is IEEE and EurAI Fellow, member of Academia Europaea and the Royal Academy of Sciences and Arts of Barcelona. She has served as Senior Editor of the IEEE Transactions on Robotics, and Associate Vice-President for Publications of the IEEE Robotics and Automation Society. Convinced that science fiction can help promote ethics in AI and robotics, one of her novels - winner of the Pedrolo and Ictineu awards - has been translated into English with the title The Vestigial Heart (MIT Press, 2018) and published together with online materials to teach a course on “Ethics in Social Robotics and AI”.
\end{IEEEbiography}

\end{document}